\documentclass[letterpaper, 10 pt, conference]{ieeeconf}  

\IEEEoverridecommandlockouts                              

\usepackage{amsmath} 
\usepackage{amssymb}  

\usepackage{xcolor}
\usepackage{graphicx}
\usepackage[subrefformat=parens,labelformat=parens]{subfig}

\newcommand{\measMat}{\ensuremath{\Phi}}
\newcommand{\basis}{\ensuremath{\Psi}}
\newcommand{\mR}{\mathbb{R}}
\newcommand{\vect}[1]{\ensuremath{\textbf{#1}}}

\title{\LARGE \bf
Compressed Learning for Tactile Object Classification
}

\author{Brayden Hollis, Stacy Patterson, and Jeff Trinkle
\thanks{B. Hollis, S. Patterson, and J. Trinkle are with the Department of Computer Science, Rensselaer Polytechnic Institute,
110 8th Street, Troy, NY, USA,\tt{ hollib@rpi.edu,\{sep,trink\}@cs.rpi.edu}.}
}

\begin{document}

\maketitle
\thispagestyle{empty}
\pagestyle{empty}

\begin{abstract}
The potential of large tactile arrays to improve robot perception for safe operation in human-dominated environments and of high-resolution tactile arrays to enable human-level dexterous manipulation is well accepted. However, the increase in the number of tactile sensing elements introduces challenges including wiring complexity, power consumption, and data processing. To help address these challenges, we previously developed a tactile sensing technique based compressed sensing that reduces hardware complexity and data transmission, while allowing accurate reconstruction of the full-resolution signal. In this paper, we apply tactile compressed sensing to the problem of object classification. Specifically, we perform object classification on the compressed tactile data. We evaluate our method using BubbleTouch, our tactile array simulator.  Our results show our approach achieves high classification accuracy, even with compression factors up to 64.
\end{abstract}

\section{INTRODUCTION}

For robots to reliably and safely function in unstructured environments, they need to perceive and react to the environment.  Sensors that sense a robot's surroundings from a distance, such as vision sensors, are very useful and commonly employed, but when it comes to physical interactions, these sensors only offer a limited perspective of what is happening.  Tactile sensors, sensors that sense the world through direct contact (e.g., force and temperature), greatly increase a robot's understanding of its interactions.  The concept of putting tactile sensors on a large portion of robots' surfaces, which we refer to as \emph{tactile skins}, has become its own research area.

There are numerous challenges with fully realizing tactile skins.  To cover the various surfaces at useful resolutions requires a large number of individual sensing elements, called \emph{taxels}, on the order of 1,000's to 1,000,000's~\cite{dahiya10tactile}.  Furthermore,  the information needs to be gathered and processed at high rates, up to $1kHz$ for fine force control~\cite{dahiya10tactile}. In conflict with these desirable features, there is limited room for the wiring of these systems, especially for skins designed separately from the robot as an add-on. In addition, tactile sensors tend to be noisy, in part from contact with the environment causing misalignments.

Even with these challenges, there are a number of potential benefits that make tactile skins worth pursuing.  One application of interest is tactile object recognition.  Often object recognition can be performed with other sensors, such as vision, but this is not always practical or possible.  For instance, a robot trying to pull items out of a crowded cupboard may not be able to get a clear view of the item of interest due to the shelf being above or below the robot's vision system or other objects being in the way.  Using tactile sensors and tactile object recognition, the robot could find the object and retrieve it with less manipulation of the environment than grabbing items one at a time for performing visual object recognition.  Another example is a rescue robot.  In some disasters scenarios, visibility is limited due to smoke or dust.  A rescue robot equipped with tactile sensors could still function and navigate using tactile object recognition to identify important landmarks, such as door knobs, or determine whether objects are movable or fixed, as was done by Bhattacharjee et al. in \cite{Bhattacharjee2012}. 

In previous work~\cite{Hollis2016}, we proposed using compressed sensing for data acquisition in tactile skins.  Compressed sensing simultaneously samples and compresses signals.  Under appropriate assumptions, the full signal can be recovered exactly or near-exactly from the compressed signal.  Our approach reduces the amount of data that needs to be gathered and transferred from the tactile skin, with little to no loss of signal resolution.  Fewer measurements imply increases in the signal acquisition rate, and,
by performing compression in hardware, the amount of wiring can be reduced.  Further, our experiments show that the reconstructed
signal exhibits less noise than the raw data sampled from the noisy taxels.

In this paper, we demonstrate additional benefits of compressed sensing for tactile skins by using the compressed signals for tactile object classification.  Object classification is a sub-problem of object recognition where objects are classified using a finite candidate set. We use a soft-margin support vector machine (SVM) to classify objects from their compressed tactile signals.  Our tactile signals are compressed from snapshots of a tactile skin with a square array of taxels pressed onto the objects from above, similar to what might happen in the cupboard example. 
Direct use of the compressed signals for classification reduces processing time because the signal reconstruction phase is omitted.
Further, it lowers the dimensionality of signals used for classification, which reduces processing time for both training  
and using the classifier.  Finally, for applications where the original signal may be required after classification is performed, for example
in grasping, the full tactile signals can be recovered it from the compressed signals.

Tactile object classification is an active area of research.  
A number of works manipulate or make multiple contacts with the objects while gathering the tactile data~\cite{Russell2000,Heidemann2004,Schopfer2007,Schopfer2009,Takamuku2008,Schneider2009,Gorges2010,Hosoda2010,Pezzementi2011,Schmitz2014}.  These extended interactions allow them to obtain multiple tactile perspectives of the object, which reduces uncertainty.  We utilize individual tactile snapshots because for some applications, like the cupboard example, extended manipulation is not practical due to space or time constraints.  Additionally, for a number of these cases, extended interaction is not necessary because most objects have a limited number of stable poses.

Jimenez et al. also use single tactile snapshots for their classification \cite{Jimenez1997}.  They still handle pose uncertainty, though, by pre-processing the full raw signal so all the observations are centered and oriented the same way in the tactile image.  The pre-processed data is then used in a neural network for classification.  We perform no pre-processing on our signals as this would be very challenging, if not impossible, in the compressed domain. Nonetheless, we achieve high classification accuracy.

Previous works also differ in the types of data used for classification.
Some approaches use the full raw tactile data~\cite{Jimenez1997,Hosoda2010,Schmitz2014}.  
A number of the methods reduce the dimensionality of the data before, or as part of, the machine learning process.  Some manually reduce the dimension by selecting features such as average force and contact area~\cite{Schopfer2009,Bhattacharjee2012,Drimus2014}. Others use automated methods to reduce the dimension, for instance, Self-Organizing Maps~\cite{Takamuku2008,Gorges2010} and Principle Component Analysis~\cite{Heidemann2004,Schopfer2007,Pezzementi2011}.  
In our approach, the dimension of the data is reduced as part of the acquisition process, which may be done in hardware~\cite{Duarte2011}.  This dimension reduction is not a pre-processing step.

The rest of the paper is organized as follows.  We give a general overview of SVMs in Section~\ref{background.sect}.  We review compressed sensing theory and our work in applying it to tactile skins in Section~\ref{cs.sect}.  Section~\ref{methods.sect} describes compressed learning theory and our application of it to tactile skins.  Our experiments and data are explained in Section~\ref{setup.sect} and \ref{data.sect}, respectively.  We present our results in Section ~ref{results.sect}, and in Section \ref{concl.sect}, we conclude with some final thoughts and ideas for future work.


\section{BACKGROUND} \label{background.sect}

Support Vector Machines are a popular classification tool in machine learning.  They classify the observations by finding a hyperplane that separates the training data into its two classes such that the distance between the hyperplane and the closest observations is maximized \cite{Burges1998}.  This distance is called the \emph{margin} and the closest observations are called \emph{support vectors}, as they are the  observations that  determine the classifier.  To find the hyperplane, one solves the following minimization problem,
\begin{equation} \label{svm.eq}
    \begin{aligned}
        & \underset{b,\vect{w}}{\text{minimize}}
        & & \frac{1}{2}\vect{w}^T\vect{w} \\
        & \text{subject to}
        & & \ell_i(\vect{w}^T\vect{x}_i + b) \geq 1\\
        & & &\text{for }i = 1, \dots, N,
    \end{aligned}
\end{equation}
where $N$ is the number of observations; $\vect{x}_i \in \mR^n$, for $i = 1$ to $N$, are the observations; $\ell_i \in \{-1,1\}$, for $i = 1$ to $N$, are the class labels; $\vect{w}$ is a vector orthogonal to the separating hyperplane; and $b=-\vect{w}^T\vect{x}_0$ for any point $\vect{x}_0$ on the hyperplane.  This optimization problem is a convex quadratic program, a well studied class of problems with many implemented solvers.  Once (\ref{svm.eq}) is solved for $\vect{w}^*$ and $b^*$, to classify a point $\hat{\vect{x}}$, one simply solves $\text{sign}(\vect{w}^T\hat{\vect{x}}+b)$.

Often it is not possible to completely separate the data with a hyperplane, so SVMs can been modified to allow some observations to be misclassified as follows:
\begin{equation}\label{softsvm.eq}
    \begin{aligned}
        & \underset{b,\vect{w}}{\text{minimize}}
        & & \frac{1}{2}\vect{w}^T\vect{w}  + C\sum_{i=1}^N\xi_i\\
        & \text{subject to}
        & & \ell_i(\vect{w}^T\vect{x}_i + b) \geq 1-\xi_i\\
        & & & \xi_i \geq 0\\
        & & &\text{for } i = 1, \dots, N,
    \end{aligned}
\end{equation}
where $\xi_i$ is the amount the $i$th observation violates the margin, known as the \emph{hinge loss}, and $C$ is a parameter to balance between maximizing the margin and reducing the margin violations.  This modification is known as soft-margin SVMs.

When training soft-margin SVMs, cross-validation is used to tune the parameter $C$.  Cross-validation separates the \emph{training} set into a \emph{development} set and a \emph{validation} set. The development set is used to train multiple SVMs.  All the trained models are then evaluated with the validation set.  The model that performs the best is then retrained using the full training set and is the final learned classifier.

The SVMs discussed so far are binary classifiers. Some modification is necessary for multi-class classification.  There are various extensions (for an overview see~\cite{Hsu2002}), of which we use the Direct Acyclic Graph SVM (DAGSVM).  DAGSVM trains a binary SVM for each pair of classes~\cite{Platt2000}.  During classification, an observation is classified by a binary SVM.  That observation is then classified in another binary SVM with the previously assigned class as one of the two potential classes and the other class eliminated from future consideration.  This is continued, sequentially eliminating classes from consideration, until a single class remains. The observation is classified as an element of the remaining class.  Thus for a $M$-class problem, DAGSVM trains $\frac{M(M-1)}{2}$ binary SVMs, but classifies an observation only using $M-1$ SVMs. It was found that the order in which the binary SVMs are used for classifying the observations does not matter~\cite{Platt2000}.

While maximizing classification accuracy (the percent of classes correctly labeled) is the true objective of SVMs, SVMs can also be evaluated by the expected hinge loss $H_D(\vect{w}^+)$ over the problem distribution $D$, where $\vect{w}^+\in\mR^{n+1}$ is the vector generated by concatenating $\vect{w}^*$ and $b^*$.  More formally,
\begin{equation}
    H_D(\vect{w}^+) = \mathbb{E}_D\bigg[\max\{(0,1-y({\vect{w}^*}^T\vect{x} + b^*)\}\bigg].
\end{equation}
$\mathbb{E}_D[\cdot]$ is the expectation over $D$.

\section{COMPRESSED SENSING FOR TACTILE SKINS} \label{cs.sect}
\subsection{Compressed Sensing Theory}

In compressed sensing, a signal, for example, force readings from a tactile array at a given time,
 is simultaneously measured and compressed by taking linear combinations
of the signal components.  Specifically, the \emph{compressed signal} $\vect{y} \in\mR^m$ with elements $y_i$ is obtained from
the \emph{full signal} $\vect{x} \in \mR^n$ as follows:
\begin{equation} \label{system.eq}
    \vect{y} = \measMat \vect{x},
\end{equation}
where $\measMat = [\measMat_{ij}]$ is the $m \times n$ \emph{measurement matrix}.
If $m < n$, (\ref{system.eq}) is under-determined and, in general, $\vect{x}$ cannot be recovered.  

In compressed sensing, one considers a restricted set of signals that are sparse in some representation basis.
Formally, a signal $\vect{x}$ is $k$-sparse in a representation basis $\basis \in \mR^{n\times n}$ if there is a 
$k$-sparse vector $\vect{s}$, meaning $ \vect{s}$ has at most $k$ non-zero entries, such that $\vect{x} = \basis  \vect{s}$.
Compressed sensing theory provides conditions under which such sparse vectors can be recovered from fewer than $n$ measurements.
One such condition relates to the \emph{restricted isometry property}, which is defined as follows.
\newtheorem{myDef}{Definition}
\begin{myDef} \label{RIP.def}
A matrix $A$ satisfies the $k$-\emph{restricted isometry property} ($k$-RIP) if there exists a $\delta \in (0,1)$ such that 
\begin{equation} \label{RIP.eq}
    (1-\delta)\|\vect{s}\|_2^2 \leq \|A\vect{s}\|_2^2 \leq (1+\delta)\|\vect{s}\|_2^2,
\end{equation}
holds for all $k$-sparse $\vect{s}$.
\end{myDef}

If $\vect{x}$ is $k$-sparse in a representation basis $\basis$, and the matrix $\measMat \basis$ satisfies the $2k$-RIP, then
$\vect{x}$ can be recovered exactly from $m \in O(k \log n)$ measurements~\cite{cs}. Many efficient recovery algorithms have been proposed.
Further, it has  been shown robust recovery in the presence of noise in the signal and/or measurements is feasible~\cite{cs}.

\subsection{Application to Tactile Skins} \label{cs-tactile.sec}
In previous work~\cite{Hollis2016},  we investigated the application of compressed sensing in planar tactile arrays.   For the measurement matrix, we used the Scrambled Block Hadamard Ensemble (SBHE), which was developed by Gan et al.~\cite{gan08sbhe}  for compressed sensing in the image domain.  SBHE is a partial block Hadamard transform with randomly permuted columns and can be represented as 
\begin{equation}
    \measMat_H = Q_mWP_n,
    \label{definePhi}
\end{equation}
where $W$ is a $n \times n$ block diagonal matrix with each block a $B \times B$ Hadamard matrix. $P_n$ is the permutation matrix, which randomly reorders the $n$ columns of $W$, and $Q_m$ selects $m$ rows of $WP_n$ uniformly at random. The SBHE was selected for its potential hardware implementations since it separates the sensors into disjoint measurement groups.  This allows simpler wiring, as each measurement samples only a limited number of taxels, and measurements within a group could share wiring.  Additionally, different groups could be measured in parallel.  The number of elements in a group is equal to $B$, which we set to 32 in this work.

\begin{figure}
\includegraphics[width=\linewidth]{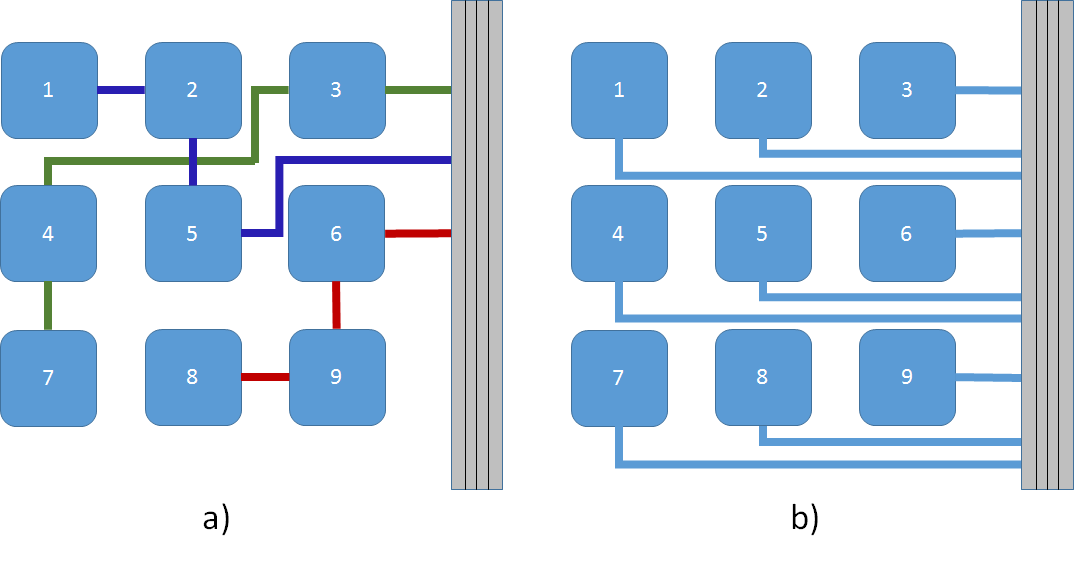}
\caption{Example wiring schematics for a) compressed sensing measurements on a $3 \times 3$ tactile grid and b) individual sensor measurements.}
\label{wiring.fig}
\end{figure}

Figure \ref{wiring.fig}(a) demonstrates the wiring on a small $3 \times 3$ array. There are three measurement groups of three sensors each.  The sensors in a group are daisy-chained together, with any measurement of that group able to use the same wire path.  
For comparison, Figure \ref{wiring.fig}(b) shows the same array with each element wired separately for individual taxel readings. 

We used the Daubechies-2 wavelet transform~\cite{daubechies88wavelet} for our basis $\basis$.  This transform is similar to the wavelet transforms used in image compression such as JPEG2000.  A similar Daubechies wavelet transform was also used in Gan et al.'s experiments~\cite{gan08sbhe}.

Using these standard (i.e. pre-existing and non-optimized) compressed sensing tools, we achieved $50Hz$ reconstruction rates for a tactile array of 4096 taxels from 1365 measurements (a compression factor of 3)~\cite{Hollis2016}. This array contains approximately the same number of sensors as the the largest existing tactile system, and the reconstruction rate is on the same order of magnitude as that system's measurement rate~\cite{icub}.  In addition, the reconstructed signal had less noise than the raw signal, and the system has potential for wire reduction.


\section{COMPRESSED LEARNING FOR TACTILE SKINS}  \label{methods.sect}

  
We propose a technique for applying compressed sensing to tactile object recognition.  Specifically, we perform object classification on the compressed signals.  This approach maintains the benefits of compressed sensing described above.  In addition, it reduces processing time by avoiding signal reconstruction, and it also reduces the dimension of the signals used for classification.

We first briefly review the theory of classification using compressed signals and then give details of our method.

\subsection{Compressed Learning} \label{cl.sect}
Calderbank et al. first proposed classification using compressed signals, a technique that they have called \emph{compressed learning}~\cite{Calderbank2009}. 
Specifically, they show that with high probability, a soft-margin SVM trained on  compressed signals has expected accuracy similar to the best linear classifier in the uncompressed data domain.  This is formalized in the following theorem.
\newtheorem{thm}{Theorem}
\begin{thm}[Thm. 3.1~\cite{Calderbank2009}] \label{cl.thm}
Let $D$ be a distribution of $k$-sparse vectors $\vect{x}_i \in \mR^n$ such that  for all $i$, ${\|\vect{x}_i\|_2 \leq R}$, where $R$ is a known upper bound. Further, assume that for each $\vect{x}_i$ there is a label $\ell_i \in \{-1,1\}$.  Let $\measMat\in\mR^{m~\times~n}$ be a measurement matrix that satisfies $2k$-RIP with constant $\delta$.  Additionally, let 
\begin{equation*}
    S_\measMat = \{(\measMat \vect{x}_1,\ell_1),...,(\measMat \vect{x}_N,\ell_N)\}
\end{equation*}
be i.i.d. labeled instances compressively sampled from $D$, and let $\vect{z}_{S_\measMat}\in\mR^m$ be the linear classifier from the soft-margin SVM trained on $S_{\measMat}$.  Finally, let $\vect{w}_0\in\mR^n$ be the best linear classifier in the uncompressed data domain with low expected hinge loss over $D$, $H_D(\vect{w}_0)$, and large margin (hence small $\|\vect{w}_0\|_2$).  Then with probability $1-2\rho$ over $S_\measMat$:
\begin{equation} \label{bound.eq}
    H_D(\vect{z}_{S_\measMat}) \leq H_D(\vect{w}_0) + O\Bigg(\|\vect{w}_0\|_2 \bigg(R^2 \delta + \frac{\log(\frac{1}{\rho})}{N}\bigg)^{\frac{1}{2}}\Bigg).
\end{equation}
\end{thm}

In (\ref{bound.eq}), $R$, $\vect{w}_0$, and $\rho$ are fixed by the problem. 
The variable $\delta$ is as defined in Definition~\ref{RIP.def} and typically increases as the amount of compression increases.  Thus, (\ref{bound.eq}) implies greater compression leads to less confidence in the classifier's accuracy, which is to be expected.  Also, as expected, the accuracy depends on training set size $N$. As $N$ increases, the accuracy of the compressed classifier approaches that of the optimal linear classifier.

Theorem~\ref{cl.thm} addresses classification with compressed signals obtained by measuring sparse signals.
If the signals are themselves not sparse, but are sparse in some orthonormal basis $\basis$, then Theorem 1 still applies, provided 
$\measMat \basis$ satisfies $2k$-RIP~\cite{Calderbank2009}.
An important point to note is that, unlike in compressed sensing, for compressed learning it is not necessary for this basis to be known.

\subsection{Application to Tactile Skins}
We propose to use compressed learning for object classification in tactile skins.  
We generate each observation as follows.
The full signal is compressed using the SBHE matrix, described in Section~\ref{cs-tactile.sec}, to generate the compressed signal of a single time instance
of contact with an object.  
More formally, for each observation, let $\vect{x}_i$ represent an $n$-vector containing the sensor readings of all  taxels (which may be noisy) at the measurement time for contact with a single object.
We generate a single compressed signal, an $m$-vector, from $\vect{x}_i$ using the SBHE, i.e., $\vect{y}_i = \measMat \vect{x}_i$.

For classification, we use the soft-margin DAGSVM described in Section~\ref{background.sect}.  We use the DAGSVM implemented in the MATLAB SVM Toolbox from University of East Anglia~\cite{Cawley2000} with a validation set to perform a grid search for the parameter $C$ in (\ref{softsvm.eq}).

Gan et al.~\cite{gan08sbhe} prove that for many basis matrices $\basis$ that have applications in image compression,
the product $\measMat \basis$ behaves like a Gaussian i.i.d. matrix. 
Further, it has been shown a Gaussian i.i.d.matrix with $m~\in~O(k \log (n/k)/\delta^2)$ rows satisfies the $k$-RIP with high probability~\cite{baraniuk2008simple}.
Therefore, by Theorem~\ref{cl.thm}, we should obtain similar classification performance on compressed signals 
as we would on the full signals.

We validate this approach through simulations, which we detail in the following sections.


\section{EXPERIMENTAL SETUP}  \label{setup.sect}

We generated tactile array data using our BubbleTouch simulator (https://github.com/bdhollis/BubbleTouch).  BubbleTouch represents taxels as rigid spheres suspended in space by spring and damper pairs; one pair per sphere.
Contact interactions between objects and taxels are assumed to be quasistatic to avoid simulation instabilities that commonly arise in the simulation of dynamic contact models.  Tactile arrays of any shape and distribution can be created simply by creating a rigid substrate body with that shape and attaching the bases of the springs to it in the desired pattern. To explore the effect of array resolution, we created eight planar square grid arrays with the same overall dimensions ($256mm$ by $256mm$), but different resolutions ranging from $4mm$ taxels in a $64 \times 64$ array, to $256mm$ taxels in an $1 \times 1$ array.  

\begin{figure}
\includegraphics[width=\linewidth]{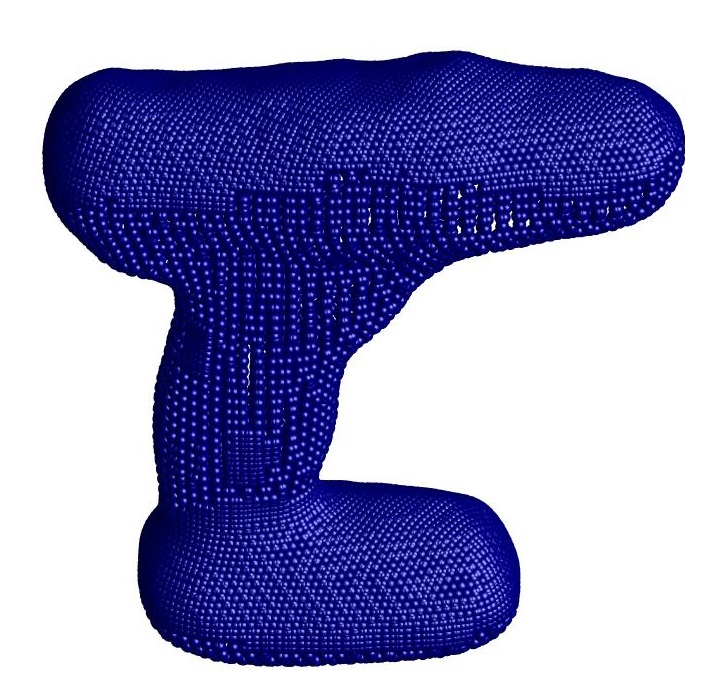}
\caption{A union-of-spheres model generated from the Yale-CMU-Berkeley (YCB) drill.}
\label{drillmodel.fig}
\end{figure}

Our objects were obtained from two sources: the YCB (Yale-CMU-Berkeley) Object Set~\cite{Calli2015} and a few simple geometric shapes.  To simplify collision detection, the objects were approximated as union of spheres. To convert a YCB object to a union-of-spheres model, we used the UC Berkley Poisson reconstructed mesh vertices as the sphere centers.  For each object, all the spheres were assigned a radius equal to twice the mean distance between all vertices and their nearest neighboring vertex.
An example of a union-of-spheres model for the YCB drill is shown in Figure \ref{drillmodel.fig}.  For primitives, for example, ellipsoids, boxes, and cylinders, we manually designed the union-of-spheres models.

\begin{figure*} 
\begin{tabular}{cccc}
\subfloat[banana (YCB)]                {\includegraphics[width = 1.5in]{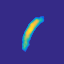}\label{banana.fig}} &
\subfloat[cup (YCB)]                   {\includegraphics[width = 1.5in]{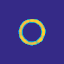}\label{cup.fig}} &
\subfloat[drill (YCB)]                 {\includegraphics[width = 1.5in]{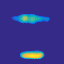}\label{drill.fig}} &
\subfloat[clamp (YCB - large)]         {\includegraphics[width = 1.5in]{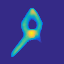}\label{clamp}}\\
\subfloat[mustard bottle - side (YCB)] {\includegraphics[width = 1.5in]{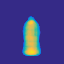}\label{mustard_side.fig}} &
\subfloat[mustard bottle - up (YCB)]   {\includegraphics[width = 1.5in]{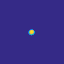}\label{mustard_up.fig}} &
\subfloat[cracker box (primitive)]     {\includegraphics[width = 1.5in]{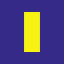}\label{cracker.fig}} &
\subfloat[cereal box (primitive)]      {\includegraphics[width = 1.5in]{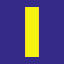}\label{cereal.fig}}\\
\subfloat[jello box (primitive)]       {\includegraphics[width = 1.5in]{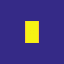}\label{jello.fig}} &
\subfloat[granola bars box (primitive)]{\includegraphics[width = 1.5in]{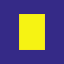}\label{granola.fig}} &
\subfloat[racquetball (primitive)]     {\includegraphics[width = 1.5in]{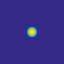}\label{racquetball.fig}} &
\subfloat[volleyball (primitive)]      {\includegraphics[width = 1.5in]{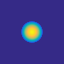}\label{volleyball.fig}}\\
\subfloat[basketball (primitive)]      {\includegraphics[width = 1.5in]{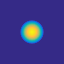}\label{basketball.fig}} &
\subfloat[gravy can (primitive)]       {\includegraphics[width = 1.5in]{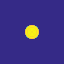}\label{gravy.fig}} &
\subfloat[tuna can (primitive)]        {\includegraphics[width = 1.5in]{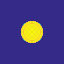}\label{tuna.fig}} &
\subfloat[salmon can (primitive)]      {\includegraphics[width = 1.5in]{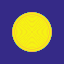}\label{salmon.fig}}
\end{tabular}
\caption{Example noiseless tactile image for each object.  In parentheses next to the object name is the object model source. The color scale goes from dark blue to yellow, where dark blue is no contact and yellow is contact of $0.01N$}
\label{objectlist.fig}
\end{figure*}


To generate tactile data, each object was placed in a stable configuration on a rigid horizontal support plane and touched from above by the tactile array.  Each touch was performed by initially positioning the substrate of the array far enough above the object avoid contact.  From there, the substrate was translated downward, causing the taxels to contact the object and move relative to the substrate (deforming the springs and dampers).  After a pre-specified downward motion of the substrate, the simulation was stopped and the spring and damper forces at all the taxels were taken as the noiseless tactile signal for this observation.  To approximate the noise in real tactile sensors, random zero-mean Gaussian noise with standard deviation of $0.001N$ (equal to $5\%$ of the signal range) was added to each taxel reading.  Values outside a taxel's range ([$0, \ 0.02$]$N$) were clipped to the boundary.  The array signal with added noise was taken as the {\em raw signal}. Figure~\ref{objectlist.fig} shows 16 images gathered from 15 objects (the mustard bottle was touched in two different orientations).  The sources of the object models are in parentheses next to their names.
%

When testing our compressed learning algorithm, we started with the raw signal from the array of the finest resolution, i.e., $64 \times 64$.  The uncompressed signal was thus of length $4,096$.  
For each raw signal, we generated a \emph{compressed signal} of length $m$ by left multiplying the raw signal by the appropriately-sized SBHE matrix. We use the following values for $m$: 1,024, 256, 64, 16, 4, and 1.

For comparison, we also performed classification using raw signals of the same dimensions as the compressed signals.  We generated raw signals using coarser tactile arrays, consisting of 1,024, 256, 64, 16, 4, and 1 taxels.
For convention, we use \emph{signal size} to refer to the number of elements in a signal's vector.  This means for compressed signals, the signal size is $m$, the number of measurements, and for the raw signals, the signal size is the number of taxels in the array.

\section{DATA SETS FOR CLASSIFICATION} \label{data.sect}
For each of the 16 objects, 360 touching observations were done as described above, but with systematic off-sets from the nominal starting configuration of the array.  The off-sets were $(0,2,4,6,8,10)mm$ along the rows of the array, $(0,2,4,6,8,10)mm$ along the columns of the array, and $(0^\circ,5^\circ,10^\circ,15^\circ,20^\circ,25^\circ,30^\circ,35^\circ,40^\circ,45^\circ)$ rotations about an axis normal to the array. This yielded 5,760 observations.  

The development set was formed by choosing $40\%$ of the 360 perturbations uniformly at random.  All observations with those 144 perturbations for all 16 objects defined the development set, which contained 2,304 observations.  The validation set was formed similarly, using $20\%$ of the other observations.  The test set was defined as the remaining 2,304 observations.  We also tested development sets using $20\%$, $10\%$, $6\%$, $2\%$, and $0.67\%$ of the 360 perturbations with the corresponding validation sets of $10\%$, $5\%$, $3\%$, $1\%$, and $0.33\%$.

\section{RESULTS} \label{results.sect}


\begin{figure}
\includegraphics[width=\linewidth]{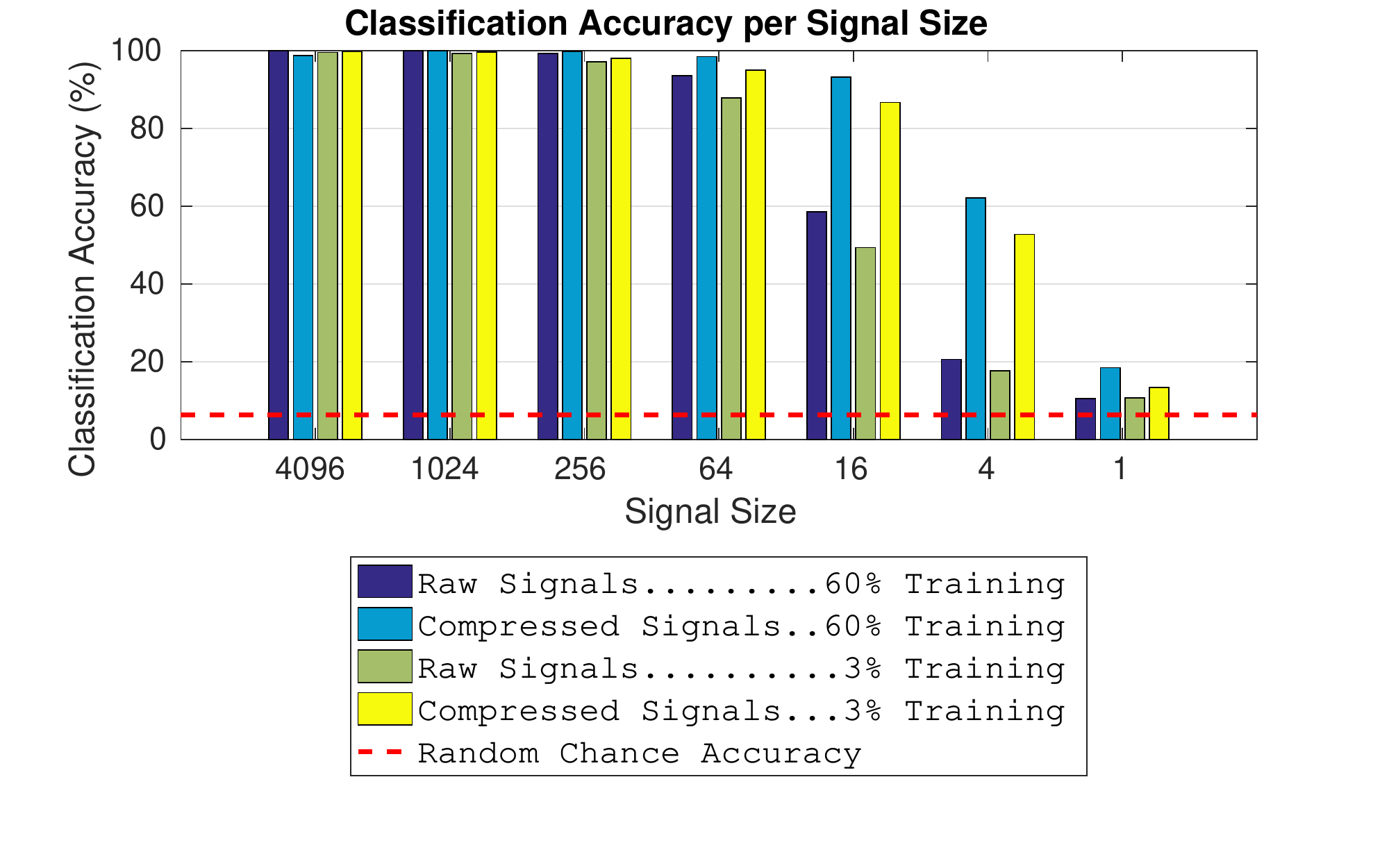}
\caption{The classification accuracy for various signal sizes using two training set sizes.  
For the raw signals, signal size refers to the number of taxels in the array, and for the compressed signals, signal size refers to the number of measurements.  The dotted line is the accuracy of randomly assigning a label to each example.}
\label{signal_size.fig}
\end{figure}

Figure \ref{signal_size.fig} shows the overall classification accuracies for seven different signal sizes. 
In addition to the accuracy rates for the compressed signals, we also look at the accuracy rates for the raw signals of corresponding dimensions. Figure~\ref{signal_size.fig} also shows the results for two different training set sizes of 60\% and 3\%.  All the results are averages over 10 different splits of the data set into test and training tests. 

Smaller signal sizes yield less accurate classification, but even with fairly small signal sizes, the classification has a high success rate. 
The raw signals achieve over 85\% accuracy even at a signal size of 64, with 93.3\% and 87.8\% for training sets of 60\% and 3\% respectively.  The compressed signals achieve that level of accuracy for the signal size of 16, a compression factor of 256, with 93.2\% and 86.7\% respective accuracies.  Overall, the compressed signals outperformed the corresponding raw signals of the same size.  The exception is for the signal size of 4,096 for which there is no compression.  
These results agree with Theorem \ref{cl.thm}.  The compressed signals have similar accuracies to the classifier on the original signal (the raw signal of size 4096).  Further, the accuracy deviates more with increased compression.

An interesting item to note is compressed signals of size less than 128 do not use every taxel value, and for the signals of size 64, at most half of the taxels are actually used.  Compressed signals of larger size  also may not involve every taxel.  

\begin{figure}
\includegraphics[width=\linewidth]{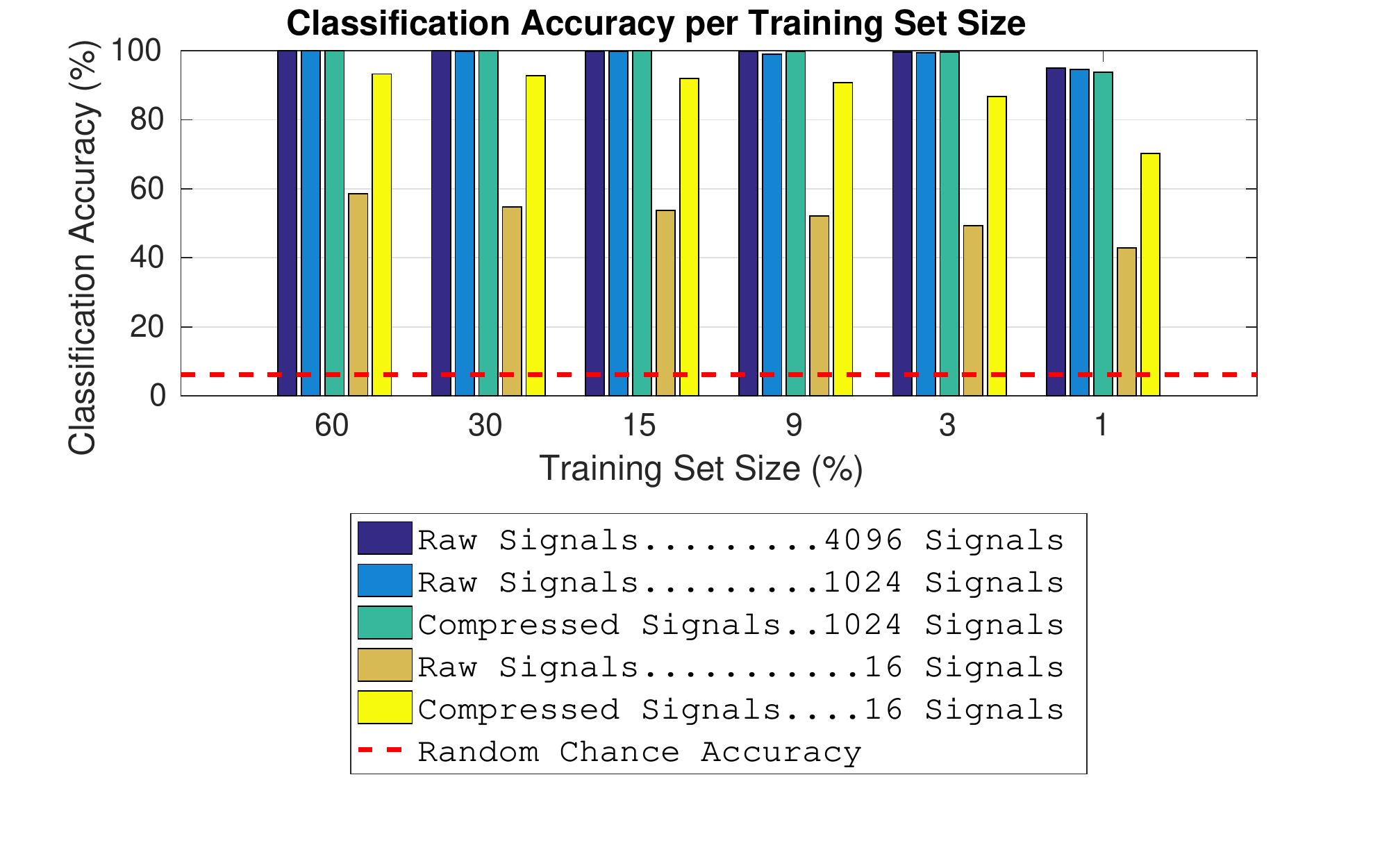}
\caption{The classification accuracy for training sets of various sizes.  
For the raw signals, the number of signals refers to the number of taxels in the array, and for the compressed signals, the number of signals refers to the number of measurements.
The dotted line is the accuracy of randomly assigning a label to each example.}
\label{training_size.fig}
\end{figure}

Collecting large sets of training data can be inconvenient, challenging, or impractical in deployed hardware.   
So, we explored performance with respect to various amounts of training data to determine how much is needed for accurate classification.  Our results are shown in Figure \ref{training_size.fig}.  It shows the classification rates for the raw signals of size 4,096, compressed signals
obtained from this raw signal, of sizes 1,024 and 16, and raw signals obtained from coarser tactile arrays of sizes 1,024 and 16.  
Again, the results are averaged over 10 splits of the data set.

The three larger signals all perform extremely well, with near 100\% accuracy for all but the smallest training set size of 1\%.  Even at the 1\% training set size, the three signals achieve over 90\% success.  For the smaller signals, the accuracy decreases as the amount of training data decreases.  The raw signals of size 16 have under 60\% accuracy for all training sizes.  The compressed signals of size 16 perform much better, maintaining over 80\% accuracy even with 3\% training data.  At 1\% training data, the smaller-sized compressed signal classification accuracy drops to approximately 70\%.

As was the case for the results shown in Figure \ref{signal_size.fig}, these results also agree with compressed learning theory.  The classification accuracies for the compressed signals continue to be similar to the results from the original signal, but the deviation increases as the training set size decreases in accordance with (\ref{bound.eq}).  This is most clearly seen in Figure \ref{training_size.fig} by comparing the deviations between original signal (raw signal of size 4,096) and the compressed signal of size 16.  The deviation goes from approximately 7\% to 15\% between 60\% and 1\% training set sizes.

\begin{figure}
\includegraphics[width=\linewidth]{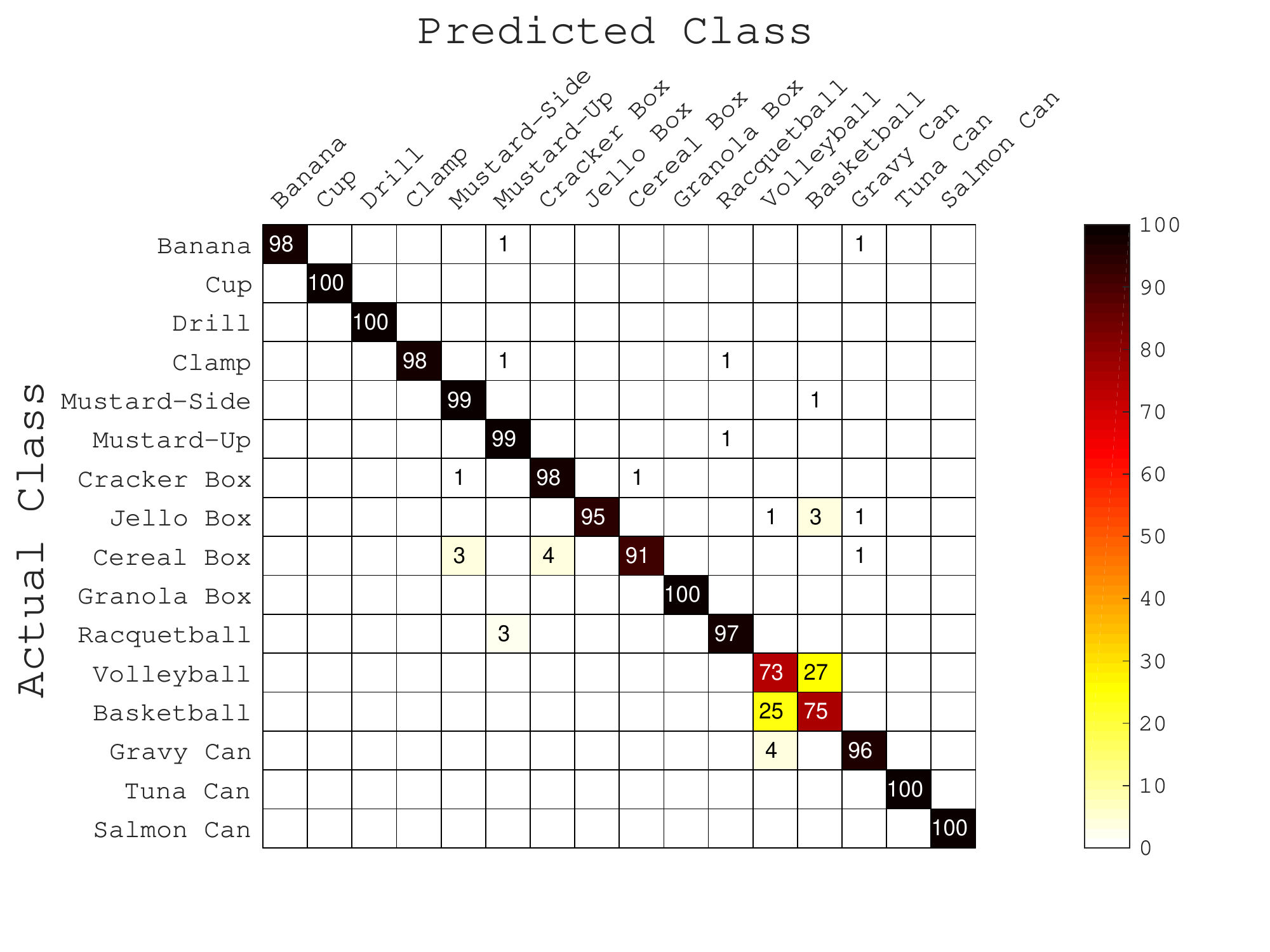}
\caption{The average confusion matrix over the 10 data-set splits for compressed signals of 64 elements trained on three percent of the examples of each object.  The values are the percentage of the actual class examples that were label as the predicted class.  Locations with no stated values have approximately zero percent of the actual class labeled as the predicted class.}
\label{confusion.fig}
\end{figure}

To get a better understanding of how the classifier is performing between individual classes, we computed the confusion matrix, which shows the percentage of observations of each class that are labeled as a particular class.  This helps to identify which pairs of classes are hard to discriminate.  Figure \ref{confusion.fig} shows the confusion matrix for the compressed signals of size 64, trained on 3\% of the observations per object, averaged over the ten splits of the data-set.  From the strong diagonal it is clear the classification performs well overall. 
The greatest confusion occurs between the volleyball and the basketball; approximately 25\% of the time one is mistaken for the other.  This is understandable because both are spheres with similar radii and similar tactile signals, as seen in Figures \subref*{volleyball.fig} and \subref*{basketball.fig}.  The gravy can and the volleyball also generate a bit of confusion.  While this is less intuitive, it is not surprising.  Both objects have round shapes, and while the volleyball has a much larger radius overall, Figures \subref*{volleyball.fig} and \subref*{gravy.fig} show the contact radii are similar.  There is also a little confusion between the upright mustard bottle and the racquetball since they also have circular contacts of similar radii.  The other confusions of note is classifying the cereal box as either the cracker box or the mustard bottle on its side and the jello box as the basketball.  This is a little less apparent, but the shape and dimensions are similar between the cereal box and the other two items, and basketball covers similar area as the jello box.





\section{CONCLUSION} \label{concl.sect}

We have developed and demonstrated an approach for tactile object classification using compressed learning.  Our approach classified various objects with high accuracy, even with high levels of compression and small amounts of training data.  The compressed signals generally resulted in performance similar to the full raw signal of individual taxel readings and outperformed raw signals of corresponding signal size.  

Our approach offers benefits of reduced data acquisition and processing time, as well as the potential to reduce wiring complexity in hardware implementations.
In addition, for tasks where the full raw signal is required after classification, this signal can be recovered from the compressed signal.
 
In future work, we will explore the application of compressed sensing to other tactile tasks, such as object manipulation, safe interaction with humans, and robot locomotion through rough terrain. In addition, we will investigate other compression techniques and wiring configurations.
Finally, we plan to implement our approach in hardware to fully evaluate its benefits.

\addtolength{\textheight}{-11.5cm}   








\bibliographystyle{IEEEtran}
\bibliography{ICRA17}

\end{document}